# RU-Net for Automatic Characterization of TRISO Fuel Cross Sections

Lu Cai[a]*, Fei Xu[a]*, Min Xian[b], Yalei Tang[a], Shoukun Sun[b], and John Stempien[a]

[a] Idaho National Laboratory, 1955 N. Fremont Ave., Idaho Falls, Idaho 83415, United States

[b] University of Idaho, 1776 Science Center Drive, Idaho Falls, ID Idaho Falls, Idaho 83402, United States

*Corresponding authors: lu.cai@inl.gov

fei.xu@inl.gov

**Abstract**

During irradiation, phenomena such as kernel swelling and buffer densification may impact the performance of tristructural isotropic (TRISO) particle fuel. Post-irradiation microscopy is often used to identify these irradiation-induced morphologic changes. However, each fuel compact generally contains thousands of TRISO particles. Manually performing the work to get statistical information on these phenomena is cumbersome and subjective. To reduce the subjectivity inherent in that process and to accelerate data analysis, we used convolutional neural networks (CNNs) to automatically segment cross-sectional images of microscopic TRISO layers. CNNs are a class of machine-learning algorithms specifically designed for processing structured grid data. They have gained popularity in recent years due to their remarkable performance in various computer vision tasks, including image classification, object detection, and image segmentation. In this research, we generated a large irradiated TRISO layer dataset with more than 2,000 microscopic images of cross-sectional TRISO particles and the corresponding annotated images. Based on these annotated images, we used different CNNs to automatically segment different TRISO layers. These CNNs include RU-Net (developed in this study), as well as three existing architectures: U-Net, Residual Network (ResNet), and Attention U-Net. The preliminary results show that the model based on RU-Net performs best in terms of Intersection over Union (IoU). Using CNN models, we can expedite the analysis of TRISO particle cross sections, significantly reducing the manual labor involved and improving the objectivity of the segmentation results.

# 1. INTRODUCTION

Tristructural isotropic (TRISO) fuel is considered a promising candidate for Generation IV nuclear reactors, and improving safety, efficiency, and sustainability are the primary goals of its design and use. It was originally developed for high-temperature gas reactors (HTGRs) [1, 2], and due to its robust design it has since gained attention for possible use in microreactors [3], space reactors [4], and fluoride salt-cooled high-temperature reactors [5]. TRISO fuel particles are commonly <1 mm in diameter and typically consist of a uranium dioxide ($UO_2$) or uranium-oxycarbide (UCO) fuel kernel that is coated with a buffer layer, an inner pyrolytic carbon (IPyC) layer, a silicon carbide (SiC) layer, and an outer pyrolytic carbon (OPyC) layer. The SiC layer serves as the primary structural barrier to retain fission products, while the IPyC and OPyC layers support and protect the SiC and act as additional barriers against the release of fission products (FP), especially fission gases. The buffer layer, a typical a porous carbon layer, accommodates kernel swelling (due to the accumulation of FP) and the buildup of gaseous FP. Each layer plays a critical role in ensuring that the fuel can withstand high-temperature and prolonged irradiation conditions, while effectively retaining FP.

Typically, TRISO particles are overcoated in a carbonaceous material referred to as the "matrix," and then thousands of the overcoated particles are formed into cylindrical fuel "compacts" or spherical fuel "pebbles." Historically, compacts have been nominally 25 mm long and 12.5 mm in diameter, and fuel pebbles have been 60 mm in diameter. Depending on the design, a TRISO-fueled reactor may have tens of thousands of compacts or pebbles in its core.

Detailed microstructural analysis of the kernels and individual layers of TRISO particles in both the as-fabricated and post-irradiation states is used to understand fuel performance. Details of interest in fuel performance include the evolution of fuel morphology as a result of irradiation-induced dimensional changes and thermomechanical interactions between the TRISO coatings, chemical interactions between FP and TRISO coatings, coating integrity, and FP retention. In the as-fabricated state, defects and deviations from desired specifications (such as layer thickness) can be identified to ensure only high-quality fuel particles are selected for use. During post-irradiation examination, researchers need to examine the microstructural features of the kernels and TRISO layers from many different particles to gather statistically significant insights or possibly observe rare phenomena. For instance, under neutron irradiation, if there is a strong interfacial bond between the buffer and IPyC layer, buffer densification can pull the IPyC and contribute to IPyC fracture. This fracture can create a path for FP (e.g., palladium [Pd]) to attack the SiC, potentially causing SiC failure that would allow the release of some FP such as cesium (Cs) [6]. Conversely, weak bonding between the IPyC and buffer layers can result in delamination (typically via fracture just barely into the buffer side of the buffer-IPyC bond), creating a gap with low thermal conductivity (occupied by FP, especially fission gases) that can increase the kernel's temperature [7]. Understanding what irradiation-induced phenomena occur, under what irradiation conditions (e.g., neutron fluence and irradiation temperature) they occur, and how often they occur helps establish suitable operational envelopes

(e.g., temperature and burnup limits) for TRISO fuel in nuclear reactors. This may also suggest areas for optimizing the design of future TRISO-based fuels.

While automated analysis of optical microscopic images of as-fabricated TRISO particles, developed at Oak Ridge National Laboratory, has significantly improved the qualification of key parameters like layer thickness for quality control [8], detailed microstructural analysis on post-irradiated particles remains predominantly manual due to the complexities induced by irradiation [9-12]. Since each fuel compact contains thousands of TRISO particles, manually obtaining statistical information on these phenomena through detailed microscopic analysis is labor-intensive and cumbersome. For instance, to get such information on kernel swelling, buffer densification, and dimensional changes of other layers, researchers have had to manually select 16 [9] or 100 points [10, 11] on each layer boundary of a ceramographic image to determine the perimeter for thousands of TRISO particles. Moreover, the challenges posed by irradiation, such as the blurring of lines between the kernel and buffer layer, can make even manual segmentation difficult and introduce subjectivity. As another example, to quantify buffer fracture frequency, researchers manually examined 1,049 particles from 13 fuel compacts [13]. Automating this examination would save researchers significant time and effort and reduce the effects of subjectivity. This study uses machine learning to expedite the process of ceramographic analysis, with the first crucial step being the automatic segmentation of TRISO fuel layers.

Segmenting TRISO fuel layers presents several challenges. First, machine-learning models require extensive datasets to develop robust algorithms, yet there is a scarcity of annotated datasets to train the models with. Fortunately, as part of the U.S. Advanced Gas Reactor (AGR) Fuel Development and Qualification Program, thousands of irradiated particles cross sections were imaged, and their kernel, outer buffer, inner IPyC, and outer SiC boundaries were manually labeled to study the post-irradiation dimensional changes [9-11]. This database provided an unprecedented opportunity to train a machine-learning model. In this study, more than 2,000 cross-section images from AGR-2 [14], the second irradiation experiment in the program, were used for this purpose.

Second, TRISO particles have a multilayered structure with varying material properties, inconsistencies in image resolution, contrast, and noise as well as the quality of polishing complicate the segmentation process. Machine-learning techniques, such as deep convolutional neural networks (DCNNs) [15] and the context-ensembled refinement network (CERNet) [16], have been used to segment the layers of cross-section images of as-fabricated (unirradiated) TRISO particles both loose and those formed into compacts and pebbles to measure layer thickness for quality control purposes. However, segmenting irradiated TRISO particles is more challenging due to complex microstructural features, irradiation-induced damage, and the presence of fission products. Defects, such as cracks or voids, along with anomalies within the layers complicate accurate segmentation. Despite these challenges, various machine-learning models, especially convolutional neural networks (CNNs), have been successfully applied to other irradiated fuels to segment microstructural features, such as pores caused by fission gas release [17-19]. This has increased confidence in the use of CNNs for segmenting irradiated TRISO particles.

Machine-learning models, particularly CNNs, offer several advantages for addressing the challenges of TRISO layer segmentation. The advanced CNN architecture effectively handles the complex microstructures of TRISO particles, accurately distinguishing between their different layers and identifying defects. By learning to recognize subtle features and patterns that may not be easily discernible through manual inspection, CNNs enhance image analysis and minimize human error and subjectivity.

Several CNN models have been used for image segmentation tasks across various domains, demonstrating their potential for TRISO layer segmentation. The U-Net architecture, widely used for biomedical image segmentation, consists of an encoder-decoder structure with skip connections [20]. The encoder path consists of convolutional and max-pooling layers that progressively reduce the spatial dimensions while increasing the depth and capturing hierarchical features. The decoder path consists of up-convolutional layers that restore the spatial dimensions and combine the high-resolution features from the encoder path through skip connections. This combination allows for precise localization and segmentation. The Residual Network (ResNet) architecture introduces residual learning to ease the training of deep networks by employing skip connections that allow gradients to flow through the network more easily. In this way it achieves state-of-the-art results in image classification tasks [21]. The Attention U-Net architecture extends U-Net by incorporating attention mechanisms that allow the network to focus on the most relevant parts of the image, effectively highlighting the important features for segmentation. The attention gates are integrated into skip connections, allowing the model to suppress irrelevant regions and emphasize significant features. This approach enhances the segmentation performance, especially in complex structures such as TRISO fuel particles, where certain layers may have subtle differences that need to be captured accurately [22]. Inspired by ResNet architecture, Residual U-Net combines the U-Net architecture with residual connections, [23]. Residual connections help mitigate the vanishing gradient problem by allowing the network to learn to identity mappings, which facilitates the training of deeper networks. The architecture retains the encoder-decoder structure with skip connections from U-Net but integrates residual blocks that improve the model's ability to learn and generalize from the data. Shareef et al. [24] proposed ESTAN (Enhanced Small Tumor-Aware Network), a new model that uses two encoders to capture image context at different scales and row-column-wise filters to adapt to the anatomy of breasts on ultrasound images.

In this research, we utilized three existing CNN architectures—U-Net, ResNet, Attention U-Net—and developed a new model called RU-Net, which is inspired by the ESTAN model to tackle the TRISO layer segmentation challenge. Each model brought unique architectural features and strengths to the task of accurately segmenting the multilayered structure of TRISO fuel particles. Specifically, the proposed RU-Net has combined a basic encoder and a ResNet encoder to accurately detect the boundaries between different layers. By leveraging the strengths of these advanced CNN models, we accurately and efficiently segmented TRISO fuel layers.

## 2. PROPOSED METHOD

### 2.1 Materials and Data Preparation

A Leica optical microscope was applied to produce images of layers of TRISO particles. The sample preparation and analysis methods used are detailed in References [10, 11] and are briefly summarized below. Four irradiated AGR-2 compacts were electrolytically deconsolidated to free TRISO particles from the surrounding graphite matrix. Table 1 summarizes their irradiation conditions. All the layers for the TRISO particles from Compact 5-4-2 were left intact. In contrast, the OPyC layer was removed from the particles of Compacts 3-3-1, 6-3-3, and 5-3-3, leaving the SiC layer as the outermost layer [13]. These particles were selected after undergoing oxidation in air at 750°C for 48 hours to remove the OPyC. This is a standard practice used in the course of measuring fission products during PIE. Each particle was embedded in epoxy and went through a series of iterative grinding, polishing, and imaging steps. Generally, there was a gap between the buffer and IPyC layers due to buffer densification and buffer-IPyC separation during irradiation. This gap was filled with epoxy during sample preparation, and in the discussion below it is referred to as the "epoxy" layer.

Table 1. Irradiation properties of AGR-2 compacts selected for this study [13].

| Compact | Kernel Material | With OPyC | Burnup (% FIMA[1]) | Neutron Fluence ($10^{25}$ n/m$^2$, E > 0.18 MeV) | Time-Average Volume-Average (TAVA) Temperature (°C) | Number of Available Cross-Section Images |
|---|---|---|---|---|---|---|
| 3-3-1 | UO$_2$ | N | 10.46 | 3.49 | 1,062 | 785 |
| 6-3-3 | UCO | N | 7.46 | 2.14 | 1,060 | 657 |
| 5-4-2 | UCO | Y | 12.03 | 3.14 | 1,071 | 407 |
| 5-3-3 | UCO | N | 10.07 | 2.91 | 1,093 | 322 |

1. FIMA: fissions per initial metal atom

To train and evaluate the supervised machine-learning models, we prepared a ground truth image for each sample image, in which each pixel was assigned a class label—i.e., kernel, buffer, epoxy, IPyC, SiC, OPyC, or background. Note that the OPyC layer was present only in particles from Compact 5-4-2. This process is illustrated in Figure 1. For each cross-section image, the researchers manually selected 100 points on the outer boundary of the kernel, the outer boundary of the buffer, the inner boundary of the IPyC, and the outer boundary of the SiC on the original image (Figure 1[a], *top*) to annotate these four boundaries (Figure 1[a], *bottom*). It is important to note that the inner boundary of the IPyC may contain a small amount of buffer that has detached from the majority of the buffer layer due to irradiation-induced IPyC-Buffer separation. In the current ground truth, these minor buffer remnants were ignored and treated as part of IPyC for simplicity. Future work aims to provide more detailed annotations, which is an ongoing effort. Leica Application Suite software then connected 100 user-selected points with straight-line segments to determine a perimeter (examples of the values of perimeters labeled in Figure 1[a], *bottom*). These boundaries were extracted (Figure 1[b], *top*) and filled in between to determine the respective layers. The boundary between the SiC and IPyC layers was automatically generated in the present work using intensity differences. The SiC and IPyC layers were extracted and converted into a grayscale image (Figure 1[b], *center*). In the grayscale image, each pixel has a value (intensity) from 0 to 255, and the regions of lower intensity appear darker. The SiC layer consistently has a higher-pixel intensity than the IPyC layer. The two layers were distinguished using the adaptive multithresh function in MATLAB [25] to determine a threshold value. Any intensity below the threshold value was considered IPyC, while any intensity above the threshold value was considered SiC. Labeling the OPyC layer required a manual approach, because, as shown in Figure 1(b), *bottom*, the pixel intensities of the OPyC layer were similar to the those in the background. In this work, a click-based interactive segmentation tool [26] allowed the user to extract the OPyC layer by providing positive and negative clicks on the image to identify the foreground and background regions, respectively. This approach leverages expert input to ensure the OPyC layer is precisely delineated amid the nonuniform intensity of the optical images.

The input images for deep-learning models were converted to grayscale from the original color images (Figure 1[a], *top*). The original images were either 2,560 by 1,920 pixels or 1,280 by 960 pixels. These grayscale images, along with the ground truth images, were cropped to minimize background interference (Figure 1[c]). During machine-learning model training, each image and its corresponding ground truth image were automatically resized to either 512 by 512 pixels or 256 by 256 pixels. The cropping operation ensured that each image was close to being the shape of a square to minimize distortion during resizing. The input images and the corresponding ground truth were manually verified by overlaying the ground truth image with some degree of transparency over the input image to ensure accuracy before being used in machine learning.

In summary, we generated a large dataset composed of 2,171 cross-section images of TRISO particles and their corresponding annotated images. Out of the total, 407 cross-section images have the OPyC layer, while the remaining images do not. Eight images do not show fuel kernels because the cross sections are either at the very top or bottom of the particles, where the kernels are either not visible or no longer present, respectively.

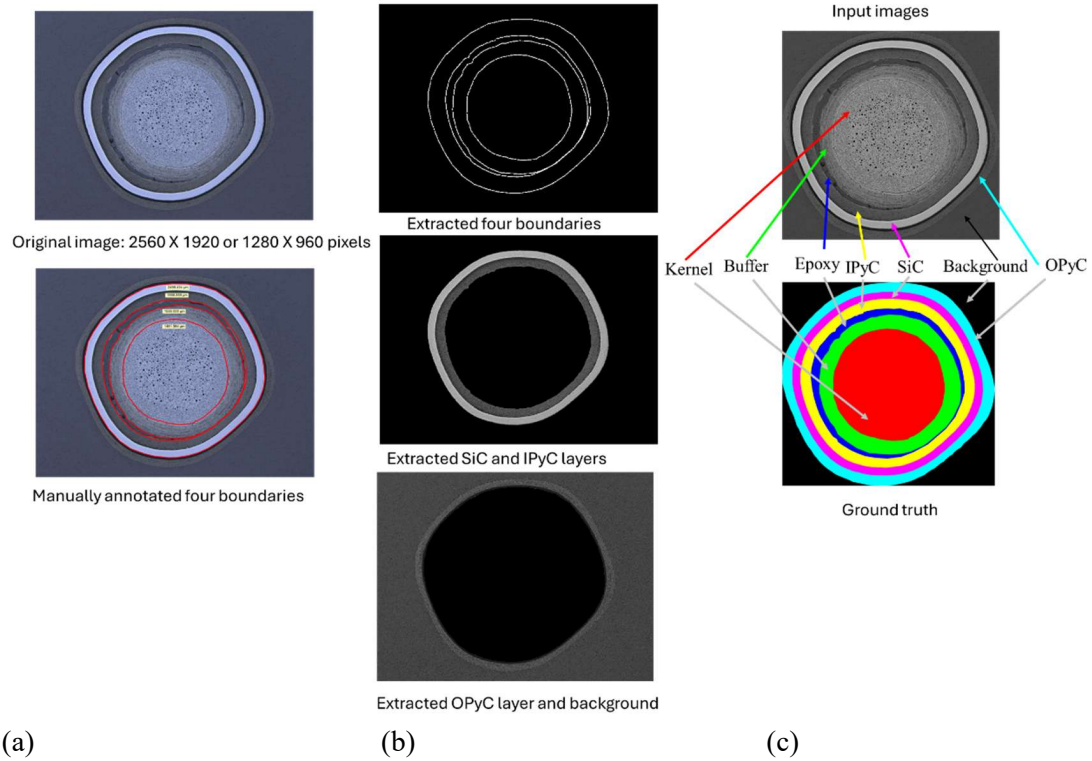

(a)                  (b)                  (c)

Figure 1. Illustration of ground truth generation: (a, *top*) An example of the original optical image and the (*bottom*) image manually annotated with red circles; (b, *top*) boundaries extracted from the red lines, (*middle*) extracted SiC and IPyC layers, and (*bottom*) extracted OPyC layer + background; (c, *top*) the input image that was cropped from the original image (a) and converted to grayscale, and the (*bottom*) ground truth indicated by different colors.

### 2.2 RU-Net for Automated Layer Segmentation

In the ESTAN model, Shareef et al. [24] introduced the use of two encoders to extract and fuse image context information at different scales, and demonstrated that the feature maps extracted from two encoders carry multiscale context information and could generate better segmentation performance for objects with different sizes. Inspired by two-encoder networks [24] that extract objects at different scales, our RU-Net architecture, which we propose for the semantic segmentation of TRISO layers, incorporates the backbone of the ResNet-50 into a U-Net architecture.

The proposed RU-Net has the basic encoder and ResNet encoder (Figure 2). The basic encoder consists of five convolutional blocks, and each block has two convolution operations followed by a max pooling.

All convolutions have a filter size of 3 by 3. The number of filters from block one to five is 32, 64, 128, 256, and 512, respectively. The two convolution layers in each block share the same setup. The ResNet encoder used the five pretrained residual blocks/stages of the ResNet-50. Starting from the second residual block, the output of the first batch normalization of the $i$th residual block was concatenated with the result of the first convolution of the $i$th encoder block (i.e., $i = 2, 3, 4$, and 5). The decoder has four upsampling blocks and one output layer. Each upsampling block has one upsampling operation followed by two convolution operations. All convolutions have the same filter size (3 by 3). The number of filters from upsampling layer one to four is 256, 128, 64, and 32, respectively. Four skip connections from the encoder and decoders in the original U-Net are used to preserve both high-level and low-level features.

We compared the performance of the existing U-Net, ResNet, and Attention U-Net with the proposed RU-Net. We leveraged existing source codes, ensuring that these models were implemented with their standard configurations for a fair comparison. For the RU-Net, we adopted the weight of the ResNet-50 model pretrained using ImageNet [27]. This approach allowed us to leverage the robust feature extraction capabilities of ResNet-50, so our RU-Net could be initialized with weights that had already learned useful representations from a broad array of images. This step is particularly beneficial in scenarios where the available dataset may not be large enough to train a deep network from scratch effectively.

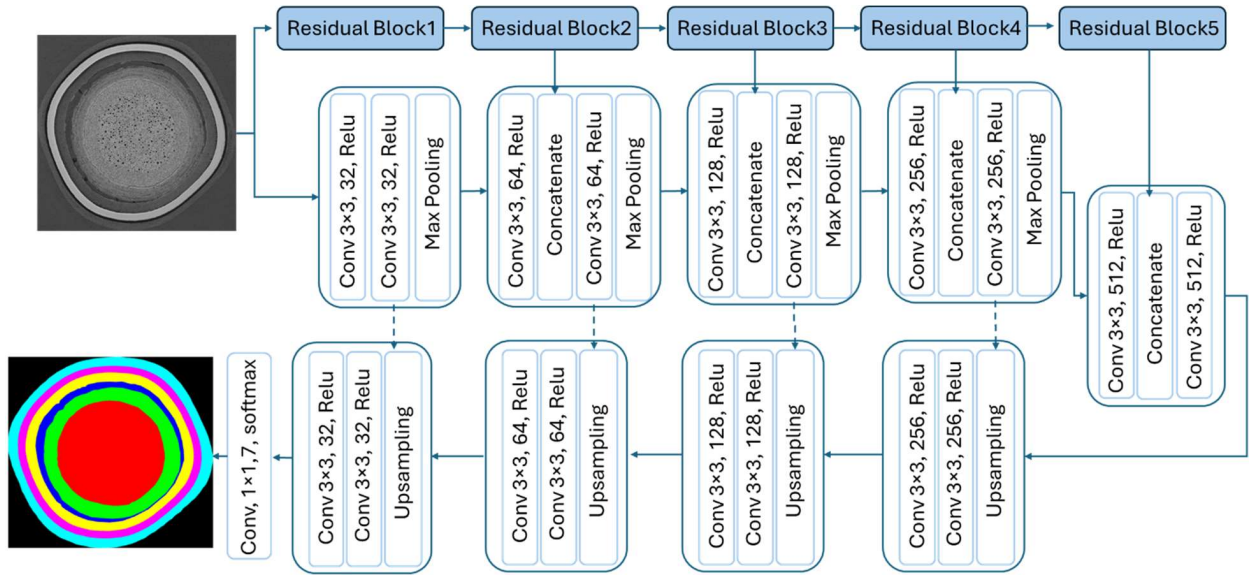

Figure 2. RU-Net model architecture. The dashed line indicates the concatenate operation.

The TRISO dataset was methodically split into three subsets: 64% was allocated for training, 16% for validation, and 20% for testing. This split ensured that the models had ample data to learn from, while also providing sufficient separate data with which to validate and test their performance, thereby reducing the risk of overfitting and ensuring a robust evaluation of model generalization. Table 2 provides detailed information about the training parameters for each model, including learning rates, batch sizes, and the number of epochs. These parameters were carefully chosen based on preliminary experiments to optimize the performance of each model. For the RU-Net, we experimented with two different image sizes: 512 by 512 pixels and 256 by 256 pixels. Training with larger images, such as 512 by 512 pixels, required significant computational resources, including Graphics Processing Units (GPUs) and the high-performance computing (HPC) infrastructure available at Idaho National Laboratory (INL). The use of these resources ensured that the model could handle the increased computational load and memory requirements associated with processing larger images. The training for the smaller images was done on a local computer. This flexibility—that it, not needing extensive computational resources—allowed us to more easily perform experimental iterations with the model. However, it is important to consider the possibility that using smaller images may impact the level of detail and accuracy in the segmentation results.

Table 2. Training parameter setup.

| Model | Training Setup |
| --- | --- |
| U-Net | Optimizer='Adam', loss = 'categorical_crossentropy', metrics= ['accuracy'], batch size = 4, learning rate = 1e-06, image size is 512×512, epochs =100 |
| Attention U-Net | Optimizer='Adam', loss = 'categorical_crossentropy', metrics= ['accuracy'], batch size = 4, learning rate = 1e-06, image size is 512×512, epochs =100 |
| ResNet-50 | Optimizer='Adam', loss = 'categorical_crossentropy', metrics= ['accuracy'], batch size = 4, learning rate = 1e-06, image size is 512×512, epochs =100 |
| RU-Net 256 | Optimizer='Adam', loss = 'categorical_crossentropy', metrics= ['accuracy'], batch size = 4, learning rate = 1e-06, image size is 256×256, epochs =100 |
| RU-Net 512 | Optimizer='Adam', loss = 'categorical_crossentropy', metrics= ['accuracy'], batch size = 4, learning rate = 1e-06, image size is 512×512, epochs =100 |

### 2.3 Evaluating the Cross-Sectional Radii and Spherical Radii

The kernels in the cross-sectional images were identified by machine-learning segmentation, and their diameters were calculated using MATLAB's "regionprops" function with the "EquivDiameter" property [25]. This property calculates the diameter of a circle with the same area as the kernels. The predicted outer diameters of the TRISO layers were also determined using the same MATLAB function, with the addition of the "imfill" function to fill the central hole of each layer. The resulting diameters, measured in pixels, were then converted to micrometers (µm) and divided by two to obtain the radii of the kernels and each layer in the cross section. These cross-sectional radii were used to calculate the spheric radii.

The mathematic model and method for calculating the spheric radii of kernels and each TRISO layer have been described in detail previously [9-11] and are summarized below. The irradiated particles are assumed to be nested spherical shells with the IPyC, SiC, and OPyC layers being concentric. The buffer and kernel are also concentric with each other but are typically not concentric with the three outer layers due to the buffer pulling away from the IPyC layer during densification under irradiation. The vertical offset between the midplanes of the outer shells and the kernel/buffer is denoted as $z_M$ (Figure 3 (e)). Cross-sectional images for each particle were obtained at four different levels or distances from the particle's mid-plane (Figure 3 (a)-(d)). Each of the four cross sections of a particle occurs at heights $z_j$ (j = 1, 2, 3, or 4) above the midplane of the outer shells. $r_i$ represents the spherical radius of the kernel and TRISO layers (Figure 3 (e)). For $i$ representing the outer OPyC, outer SiC, outer IPyC, and inner IPyC:

$$x_{ij} = \sqrt{r_i^2 - z_j^2} \quad (1)$$

For $i$ is the kernel or the outer buffer:

$$x_{ij} = \sqrt{r_i^2 - (z_j - z_M)^2}, \text{ for } |r_i| > |z_j - z_M| \quad (2)$$

where $x_{ij}$ is the cross-sectional radius of the corresponding kernel or TRISO layers, obtained from the ceramographic images. The parameters $z_M$, $r_i$, and $z_j$ are fixed for each particle but need to be solved.

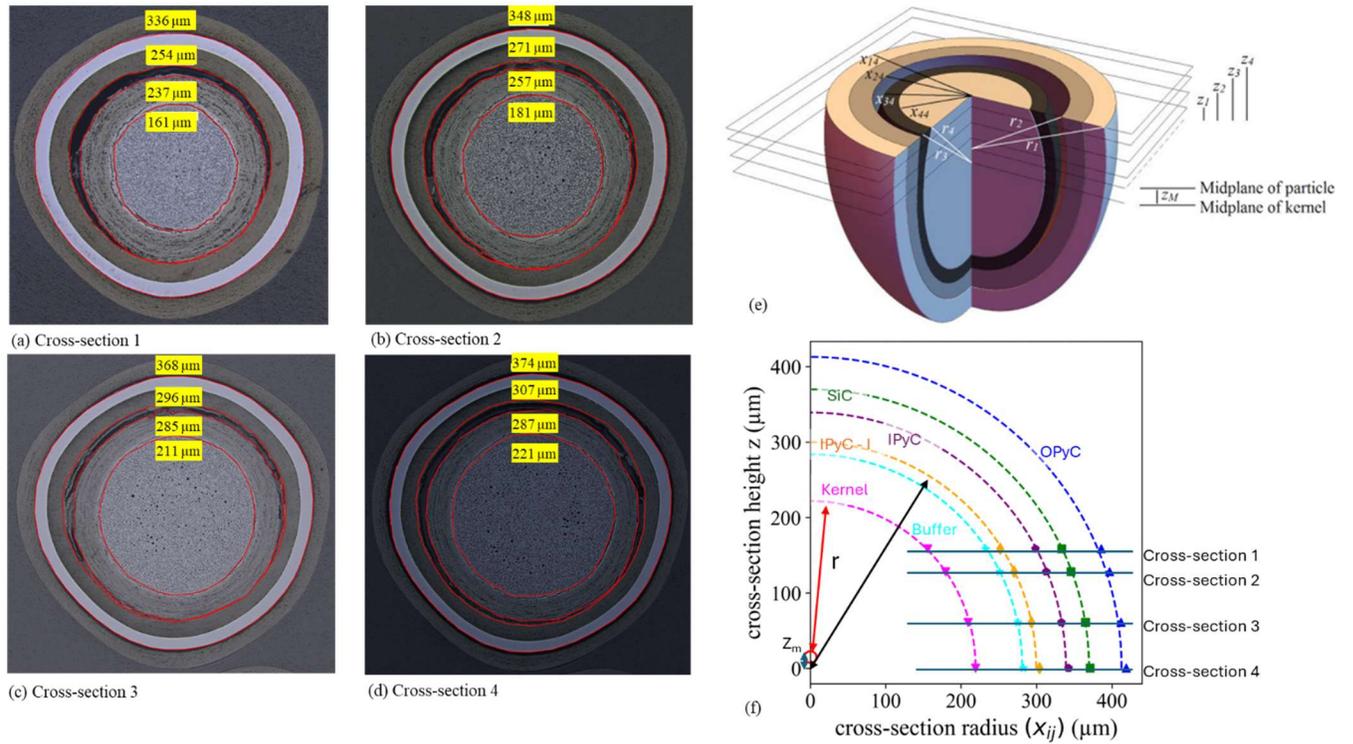

Figure 3. (a)-(d) Cross sections (with manual annotation) at four different levels within one particle from AGR-2 Compact 5-4-2 [10, 11]. The circumferences of four manually annotated boundaries are labeled in each image and the corresponding radii are also shown in the yellow rectangles. (e) An idealized particle geometry [9]. The illustration of particle geometry ($x_{ij}$ and $r_i$) includes only four boundaries: kernel, outer buffer, inner IPyC (IPyC-I), and outer SiC, for consistency with manual annotation. (f) Cross-section radii vs. cross-section height ($z_j$) for four cross-section in (a)-(d). Spherical radii (r) for kernel and TRISO layers are plotted as dashed lines. The particle center is at (0,0), while the kernel/buffer center is at (0, $z_m$). $x_{ij}$ and $r_i$ of outer OPyC and outer IPyC (or inner SiC) are also plotted in the (f), though not manual annotated in (a)-(d).

For particles without OPyC, a total of 20 radii ($x_{ij}$ for $i$ representing the outer SiC, outer IPyC, inner IPyC, outer buffer, or the kernel) were observed from the ceramographic analysis of the four cross sections per particle, while the ten parameters to be determined are the five spherical radii ($r_i$), the heights of the four cross sections from the midplane ($z_j$), and $z_M$. These parameters were determined via maximum likelihood estimation using optimization functions in MATLAB. In addition to the observed radii in the cross sections ($x_{ij}$), the radius of the silhouette, which was a shadow projected from backlighting each in a translucent mount, was measured and assumed to be close to the radius of the outermost layer, in this case, SiC. Additionally, for particles with OPyC (from Compact 5-4-2), a total of 24 $x_{ij}$ radii were observed, and eleven parameters (with additional $r_{OPyC}$) were also solved via maximum likelihood estimation. In this case, the silhouette radius served as the initial guess for the spherical radius of the OPyC for each individual particle. Figure 3 (f) plots an example of observed $x_{ij}$ values from the four cross-sections shown in Figure 3(a)-(d), against the resulting $z_j$ from maximum likelihood estimation. The spherical layers are indicated as dash lines, and $z_M$ is represented by an open dot in the y-axis.

To reduce uncertainty, only particles with complete sets of circumferences were included in this analysis, for a total of 420 particles. However, for some particles the optimization algorithm did not converge to a solution. Ultimately, spherical radii were determined for 321 particles without OPyC and 70 particles with OPyC (a total of 391 particles).

## 3. EXPERIMENTAL RESULTS AND DISCUSSION

### 3.1 Model Evaluation

The Intersection over Union (IoU), a commonly used metric, was used to evaluate and compare the performance of the different models. It is defined by

$$IoU = \frac{|P \cap G|}{|P \cup G|} \quad (3)$$

where $P$ is the predicted object region, $G$ is the ground truth region of the corresponding object, and $|\cdot|$ measures the number of nonzero pixels. Table 3 lists the IoU for each class, divided into the train and test datasets. In addition, the RU-Net model was trained on two different image sizes: 512 by 512 pixels and 256 by 256 pixels. The IoU values for the train and test datasets are consistent for each model, indicating no overfitting. In addition, the mean IoU (mIoU) for each model was calculated as the average of IoU across all classes, as shown in Table 3 [16].

Table 3. IoU comparison of different models.

| Method | mIoU | Train/Test | | | | | | |
|---|---|---|---|---|---|---|---|---|
| | | Kernel (%) | Buffer (%) | Epoxy (%) | IPyC (%) | SiC (%) | OPyC (%) | Background (%) |
| RU-Net 512 | **93.6** | **97.8/97.4** | **94.9/93.5** | **86.6/82.9** | **94.4/93.0** | 95.4/95.2 | 93.9/93.2 | **97.8/97.4** |
| Attention U-Net | 92.3 | 93.0/92.6 | 92.2/91.5 | 82.8/80.7 | 92.8/92.5 | 95.6/95.4 | 93.3/93.5 | 97.3/97.1 |
| ResNet | 90.1 | 92.9/92.7 | 89.1/88.3 | 77.7/75.7 | 91.9/91.8 | 94.4/94.2 | 90.1/89.7 | 95.7/95.4 |
| U-Net | 93.5 | 96.9/96.7 | 93.9/93.3 | 83.0/81.3 | 93.3/93.0 | 95.9/95.8 | 94.9/95.2 | 97.9/97.8 |
| RU-Net 256 | 90.6 | 91.3/91.8 | 88.6/88.1 | 79.9/77.4 | 92.5/91.5 | 95.2/94.9 | 93.2/91.0 | 97.8/97.4 |

### 3.2 Overall Performance of Layer Segmentation

As shown in Table 3, for models trained on images sized 512 by 512 pixels, the RU-Net and U-Net models performed better than the other models, though all the mIoU values exceed 90%. For the RU-Net model trained on two different image sizes, the IoU scores of the IPyC, SiC, OPyC, and background classes are consistently high, while the model trained on larger image sizes performs better in the other classes (kernel, buffer, and epoxy). The IoU for the epoxy layer is the lowest among all classes for all models. This is not surprising, as the epoxy (or IPyC-buffer gap) typically has the smallest area A slight over- or underprediction for the boundary can significantly reduce the IoU. For example, in Figure 4, there is a tiny gap between the buffer and IPyC (i.e., the epoxy class). The inner perimeter of the IPyC is 1,795.57 μm, while the outer buffer perimeter is 1,772.94 μm, as shown in Figure 4(a), corresponding to a maximum buffer-IPyC gap thickness of 7.2 μm (about 14.9 pixels in the original image or 4.5 pixels in the input image resized to 512 by 512 pixels). In addition, the radial cracks in the buffer also affect the buffer prediction. Because the purpose of the manual annotation (red circles in Figure 4[a]) was to determine the perimeter for calculating the dimension changes, the large radial cracks in the buffer were ignored. Cases with broken buffer are an important subset of the particle population, and it would be very advantageous if these radial gaps can be automatically identified through machine learning. However, this is part of ongoing work and will be addressed in a future publication. At this stage, the focus is on the layer segmentation before delving into the detailed identification of microstructural features like buffer cracks. The segmentation results from RU-Net align well with the ground truth, though the epoxy boundary is overpredicted, especially where the buffer layer shows radial cracks (Figure 4[c]). Although the visual match of the segmentation results appears good, the epoxy class has one of the lowest IoU performances at 28%. This low IoU does not necessarily indicate poor model performance.

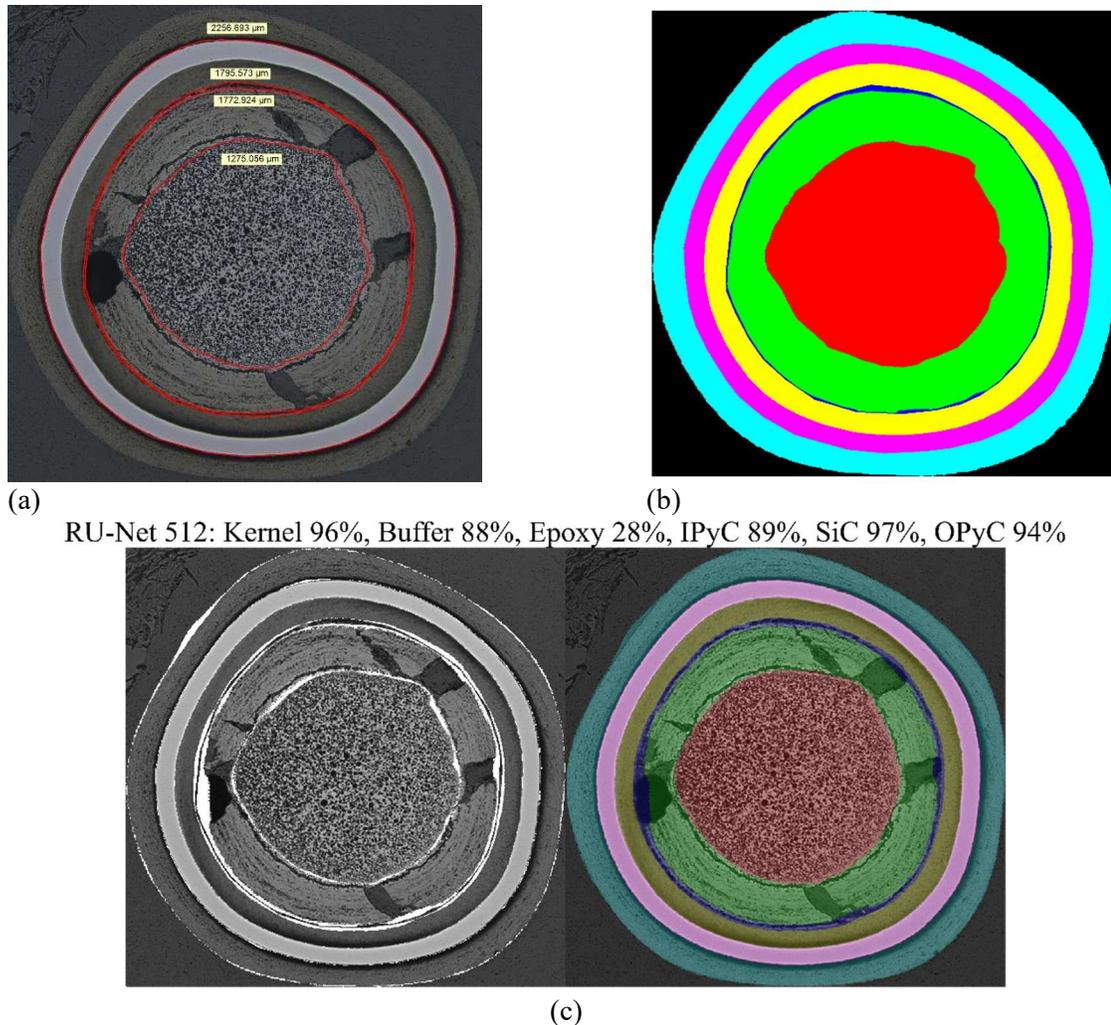

Figure 4. An example of a cross-sectional image with a small area in the epoxy class (a tiny buffer-IPyC gap): (a) manually annotated images with red circles showing different boundaries of the classes (the perimeters are shown in yellow rectangles), and the (b) ground truth. The results (c) from RU-Net 512. On the left, the input image is overlaid with the difference (shown in white) between the ground truth and the prediction. On the right the input image is overlaid with the prediction shown in different colors (red, kernel; green, buffer; blue, epoxy; yellow, IPyC; pink, SiC; and cyan, OPyC). The IoU for the kernel and TRISO layers are given above these images.

Figure 5 and Figure 6 present examples of segmentation results from different models for particle cross sections with and without OPyC. These two cross-section examples were randomly selected from the 2,171 TRISO cross-section images. The IoU for each class in these examples (shown above each pair of images) was close to the average IoU for the entire dataset, as shown in **Error! Reference source not found.**. Therefore, these two examples represent the good segmentation results that were typical of this work. The left image in each pairing is the input image overlaid with the difference (shown in white) between the ground truth and prediction. The right image for each is the input image overlaid with the prediction shown in different colors. All models segmented the SiC and IPyC well for both cross-section examples. The U-

Net and RU-Net models segmented each class accurately. The main discrepancy between the prediction and ground truth occurs at the boundaries of each class. In Figure 5(c), the ResNet model overpredicts the OPyC from the background in the ground truth along the boundary between the OPyC and the background, and in Figure 6(c) the ResNet model overpredicts the epoxy from the buffer at the right (three o'clock location). In Figure 5(d), the Attention U-Net model underpredicts OPyC in the boundary on the bottom right of the image, and in Figure 6(d) it mispredicts some scattered area inside the kernel as the buffer.

When comparing the RU-Net models trained on different image sizes, both models achieve a high IoU for the IPyC, SiC, and OPyC (if available) classes. However, the model trained on larger images performs better in the kernel and buffer classes (Figure 5[b] and Figure 6[b]), primarily at the boundaries. The smaller image size reduces resolution, increasing the challenge of segmenting precise boundaries (Figure 5[e] and Figure 6[e]).

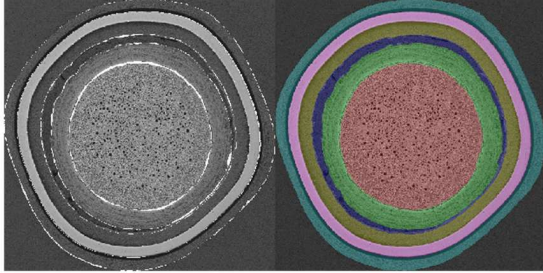
(a)

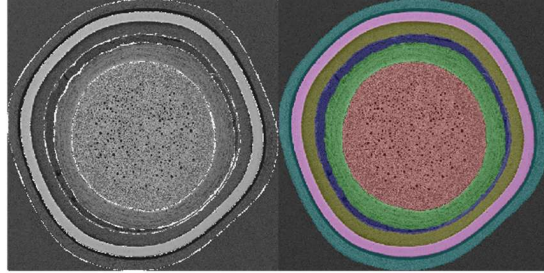
(b)

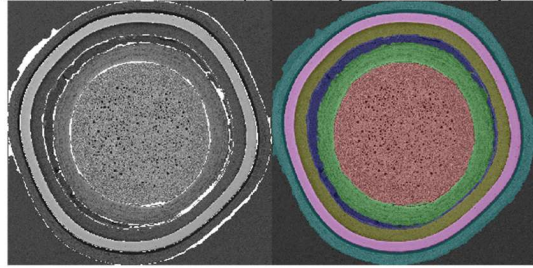
(c)

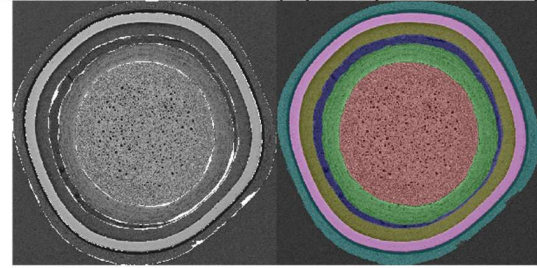
(d)

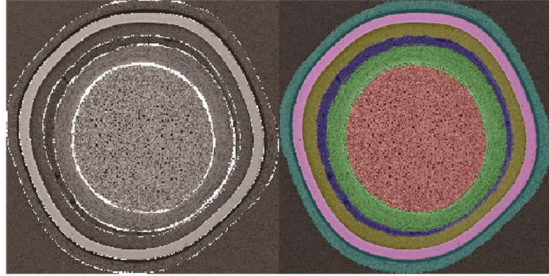
(e)

Figure 5. Evaluation results for (a) U-Net, (b) RU-Net 512, (c) ResNet, (d) Attention U-Net, and (e) RU-Net 256 for particles with OPyC. The left image in each pairing is the input image overlaid with the difference (shown in white) between the ground truth and prediction. The right image is the input image

overlaid with the prediction shown in different colors (red, kernel; green, buffer; blue, epoxy; yellow, IPyC; pink, SiC; and cyan, OPyC).

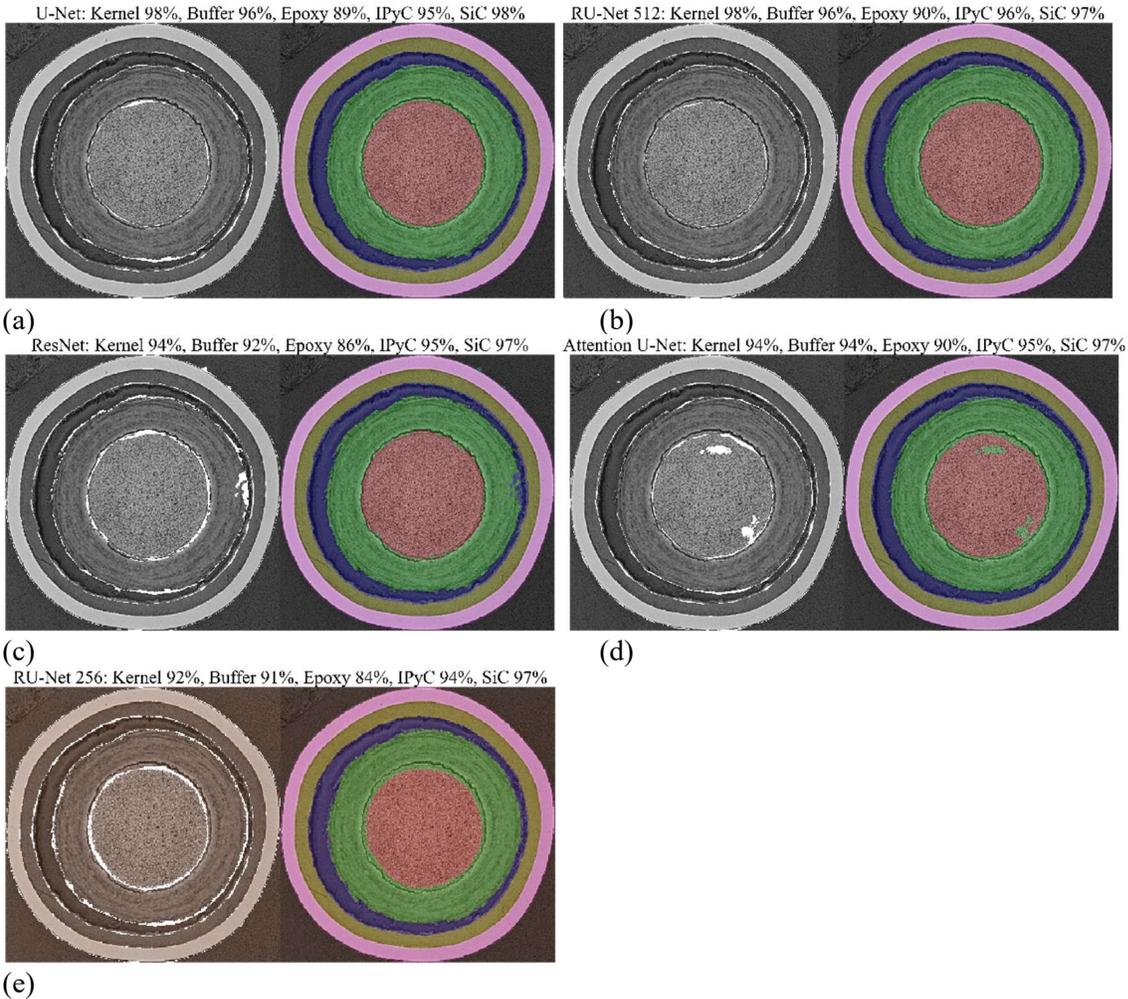

(a) (b)
(c) (d)
(e)

Figure 6. Evaluation results for (a) U-Net, (b) RU-Net 512, (c) ResNet, (d) Attention U-Net, and (e) RU-Net 256 for particles without OPyC. The left image in each pairing is the input image overlaid with the difference (shown in white) between the ground truth and prediction. The right image is the input image overlaid with the prediction shown in different colors (red, kernel; green, buffer; blue, epoxy; yellow, IPyC; and pink, SiC).

As mentioned before, to determine the spherical radii, cross-section images of each particle were obtained at four different levels. In some cases, the cross section was taken far from the midplane, or equator, of the TRISO sphere, resulting in kernels that appear small (an example is shown in Figure 7), and coatings that appear thick in-cross section. In rare cases (eight out of 2,171) the cross section does not include the kernel (see Figure 8). All models show good segmentation results for the SiC and IPyC in Figure 7. The RU-Net 512 model underpredicts the kernel (kernel IoU 25%) and the RU-Net 256 model fails to identify the kernel but segments other layers well. The U-Net model performs slightly better with the kernel area (kernel IoU 53%) but misidentifies some scattered areas inside the buffer as the kernel and significantly overpredicts the buffer from the epoxy. Similarly, the ResNet model overpredicts quite a few buffer areas as the kernel (kernel IoU only 14%) and overpredicts the buffer from the epoxy. Attention U-Net model almost fails to identify the kernel area, resulting in an extremely low IoU. It also overpredicts the buffer from the epoxy class.

In cases where the kernel is missing, both the RU-Net 512 and 256 models successfully identify the absence of the kernel and accurately segment other classes or layers. The RU-Net 512 model achieves a slightly higher IoU than the RU-Net 256 model. The Attention U-Net model also identifies that the kernel does not exist but does not segment the OPyC layer well; it misclassifies the OPyC at the eleven o'clock position in Figure 8. The U-Net and ResNet models misclassify some buffer areas as kernel. Therefore, the RU-Net 512 model performed the best in the cases of small kernel or no kernel.

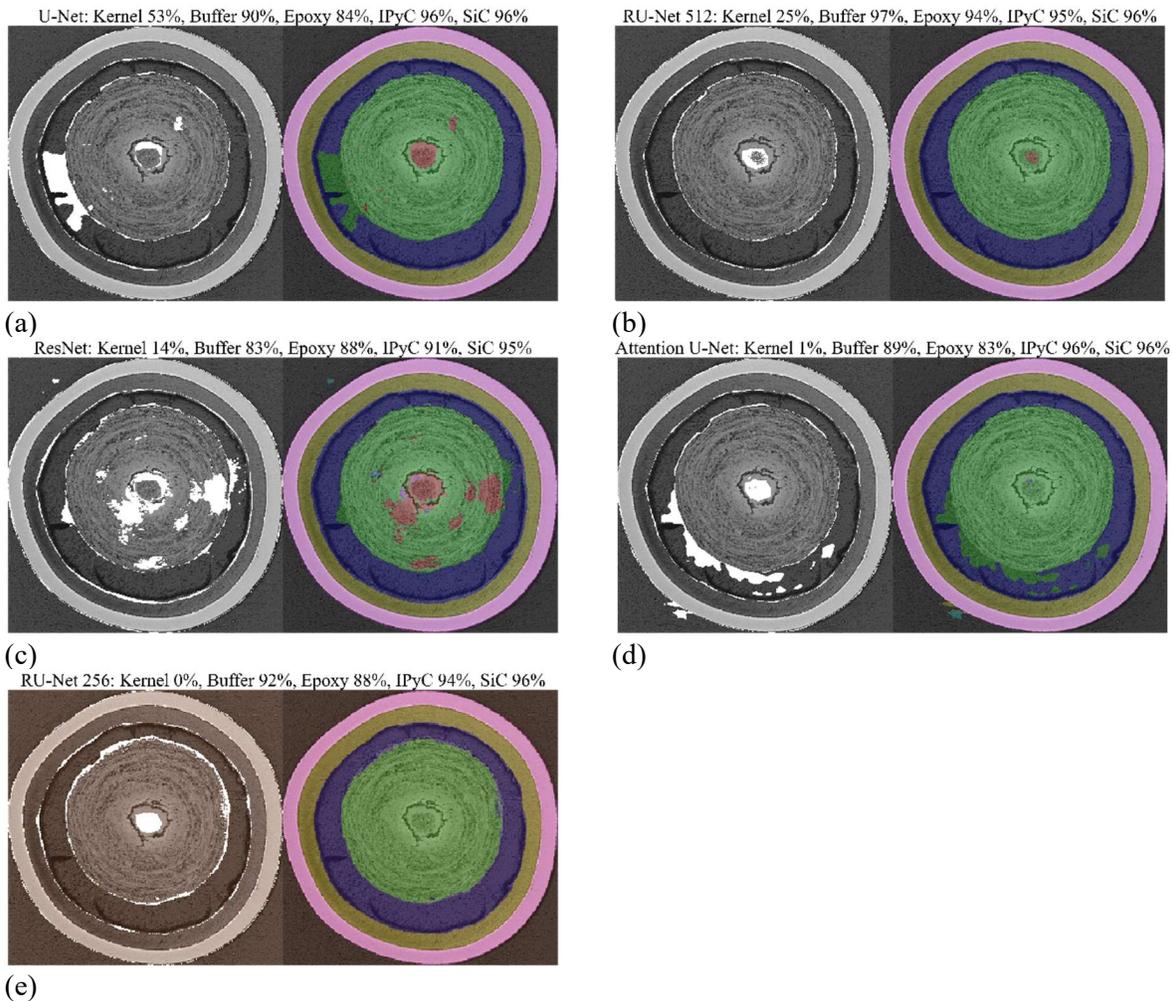

Figure 7. Evaluation results for (a) U-Net, (b) RU-Net 512, (c) ResNet, (d) Attention U-Net, and (e) RU-Net 256 for a cross section with a small kernel.

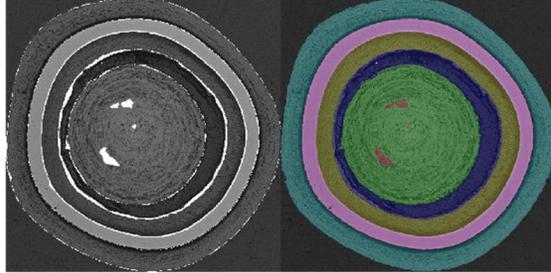
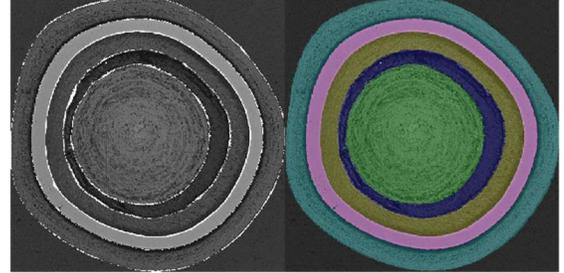
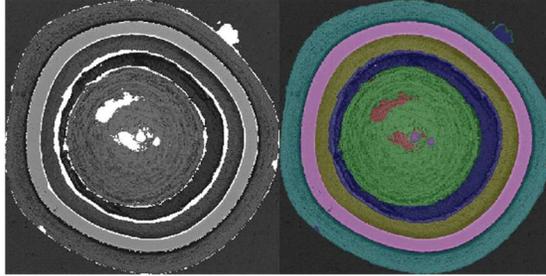
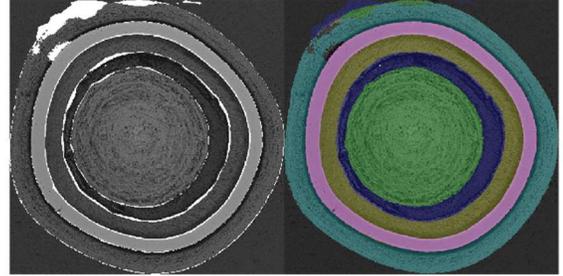
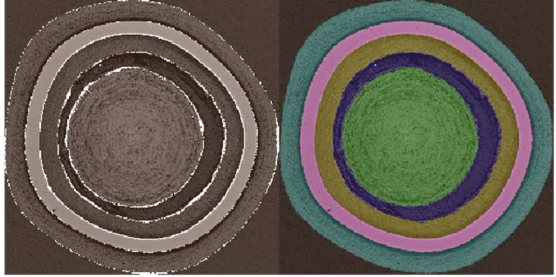

(a)  (b)  (c)  (d)  (e)

Figure 8. Evaluation results for (a) U-Net, (b) RU-Net 512, (c) ResNet, (d) Attention U-Net, and (e) RU-Net 256 for a cross section without a kernel.

It is not uncommon to observe damages or abnormalities caused by sample preparation, which can present challenges in segmentation. These issues have also been noted in cross-sectional images of as-fabricated TRISO particles, where anomalies like dirty surface and out-of-focus areas significantly skew the layer thickness measurements. As a result, a CNN model has been proposed in other work [28] to screen particle images with anomalies and exclude them from further measurement to obtain better statistics. In the current dataset, some abnormalities have also been observed, such as partially hidden layers (Figure 9[a]), layer damages (Figure 10[a]), and polishing scratches (Figure 11[a]). The sensitivity of the models to these abnormalities is examined here.

In rare cases, the IPyC layer does not appear as a continuous ring, with parts obscured at the nine o'clock position in Figure 9(a), most likely covered by the excess backpotted epoxy. Backpotting is a common practice in preparing TRISO mounts for microscopy. When the IPyC-buffer gap is accessible through grinding, additional epoxy fills these voids, then vacuum is applied to aid with ingress of epoxy into the gaps. Frequent backpotting supports the TRISO layers and prevents loss of the fuel kernels during subsequent grinding operations. The uneven texture or appearance of epoxy in the gap in Figure 9(a) indicates issues with backpotting and excess epoxy at the nine o'clock position, which hides the IPyC layer. Although this partial obscuring of the IPyC is rare, it provides a unique opportunity to test the performance of different models. The RU-Net 512 and 256 models successfully capture the fading part of the IPyC ring, as shown in Figure 9(c) and 8(f), while the other models fail to identify it, as shown in Figure 9(b), 8(d), and 8(e).

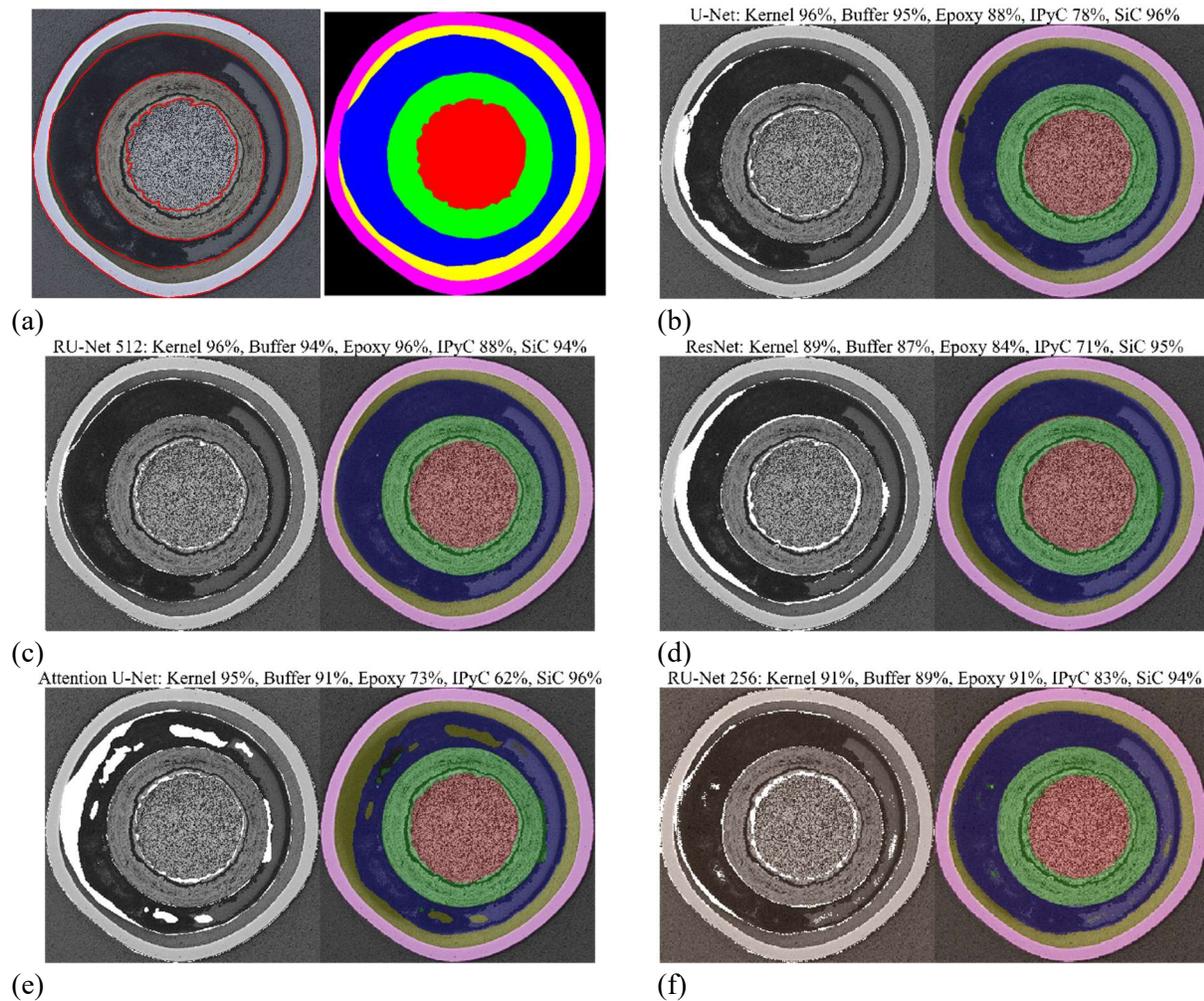

Figure 9. A cross section—(a) image (manual annotated by red circles) and ground truth—with a partially fading IPyC layer and evaluation results for (b) U-Net, (c) RU-Net 512, (d) ResNet, (e) Attention U-Net, and (f) RU-Net 256.

For cross sections with a damaged OPyC (four out of 2,171 images—i.e., four cross sections from one particle), as exemplified in Figure 10 (image and ground truth), the Attention U-Net model misidentified portions of the OPyC (around the five o'clock location) as background and incorrectly classified parts of the kernel as buffer, as shown in Figure 10(e). The ResNet model performs adequately, though it encounters some challenges at the two broken locations of the OPyC, which is shown in Figure 10(d). Figure 10(b) and 9(c) show that the U-Net and RU-Net 512 models performed well overall.

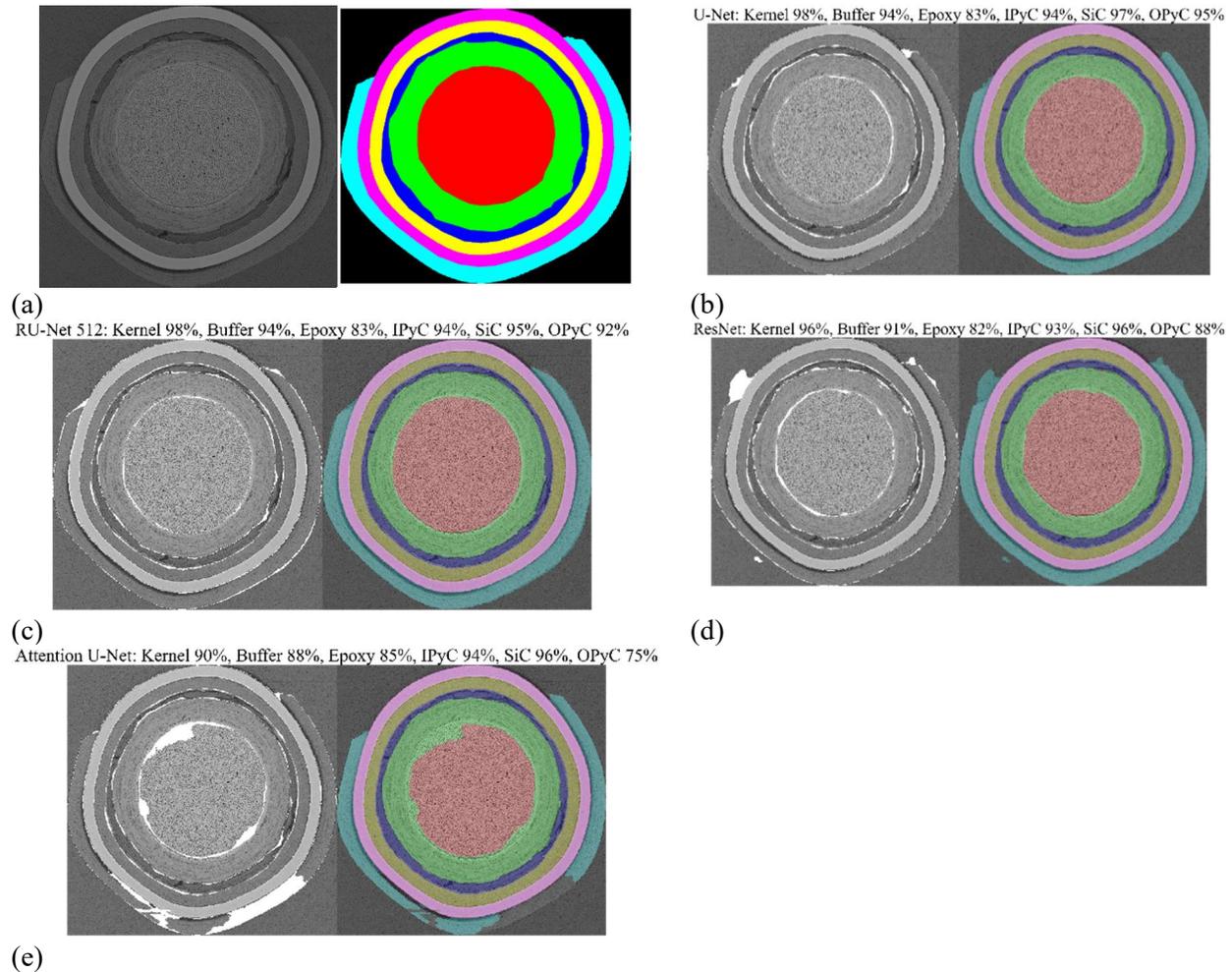

(a) (b) (c) (d) (e)

Figure 10. A cross section—(a) image and the corresponding ground truth—with a broken OPyC layer and the evaluation results for (b) U-Net, (c) RU-Net 512, (d) ResNet, and (e) Attention U-Net.

Minor artifacts such as pockmarks and scratches on the SiC layer caused by inadequate polishing are commonly observed (for example, Figure 11[a]). Both the U-Net and ResNet models successfully ignore these minor imperfections on the SiC layer, as shown in Figure 11(b) and 10(d), but the RU-Net 512 and Attention U-Net models misidentify parts of these scratches as the OPyC. For the models tested, RU-Net 512 has the best IoU values for all the classes except SiC.

To summarize this section, it is not surprising that the RU-Net model trained on images resized to 512 by 512 pixels performed better than the same model trained on images sized to 256 by 256 pixels. Training on 512-by-512-pixel images requires more computing resources but is more advantageous for detailed segmentation, especially for delineating boundaries. Using images resized to 512 by 512 pixels, the RU-Net 512 model achieved the best segmentation results among the models evaluated. It outperformed the others not only for general presentative cross sections (Figure 5 and Figure 6), but also for the extreme or rare cross sections involving small or absent kernels (Figure 7 and Figure 8) or partial obscured IPyC layer (Figure 9), and broken OPyC layer (Figure 10). However, the RU-Net 512 model does face some challenges when dealing with artifacts in the SiC remaining from sample preparation. As the results show, the RU-Net model, which combines the strengths of two encoders, is particularly beneficial for segmenting the intricate and multilayered structure of TRISO particles. Since the RU-Net 512 model performed the best, the following discussion focuses on its segmentation results.

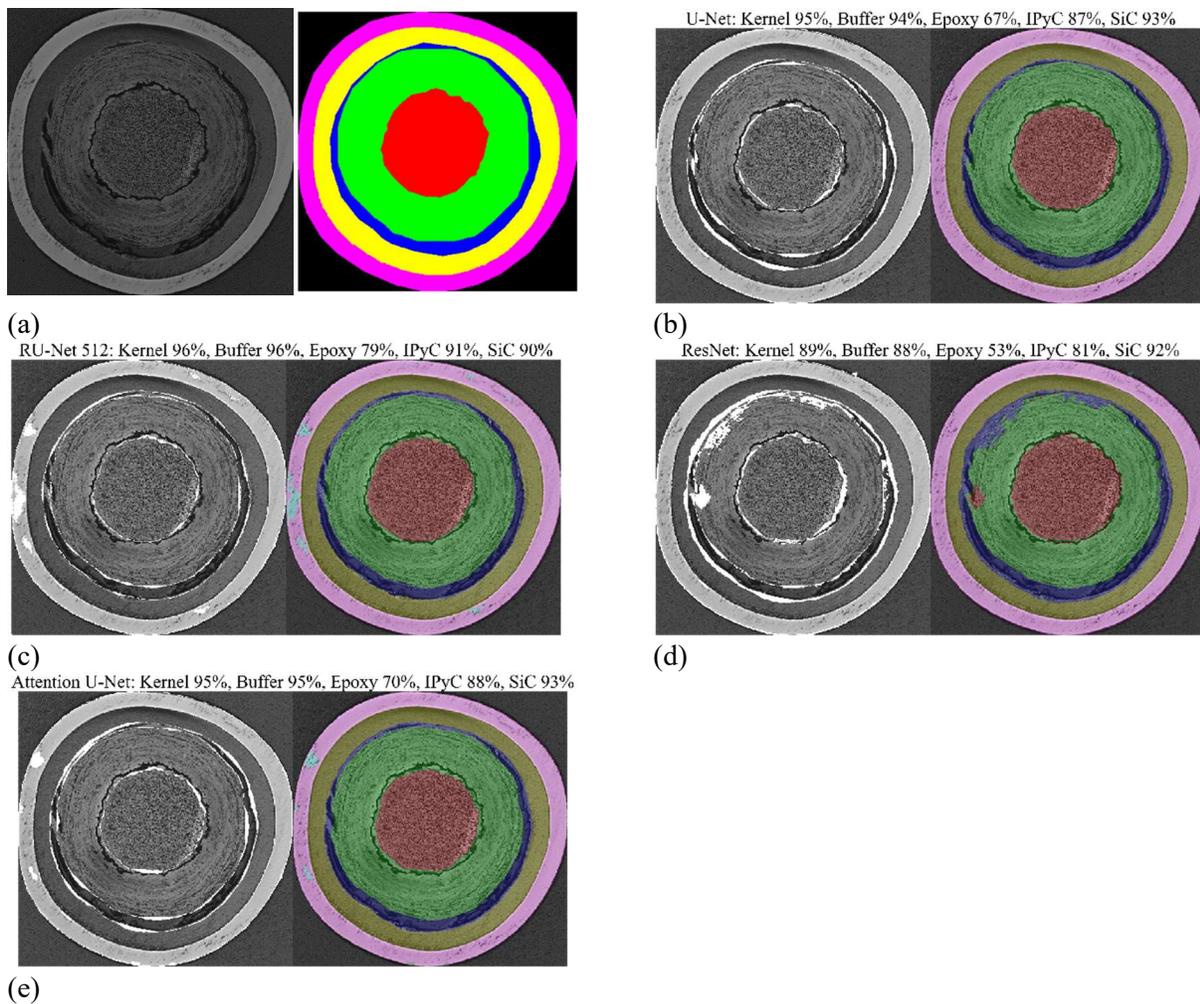

Figure 11. A cross section—(a) input image and corresponding ground truth—with polishing scratches in the SiC and the results for (b) U-Net, (c) RU-Net 512, (d) ResNet, and (e) Attention U-Net.

### 3.3 Cross-Section Radius Comparison—RU-Net 512

Since the RU-Net model with input images sized to 512 by 512 pixels outperformed the other models, this section will compare the cross-section radii from this model to those from the manual annotations. The comparison of spherical radii will be discussed in section 3.4. It is important to clarify the difference between these two types of radii: the cross-section radii are directly derived from each of the 2,171 cross-sectional images, while the spherical radii are determined via maximum likelihood estimation from the four cross-sectional images of each particle, as described in section 2.3. This section focuses exclusively on the discussion of cross-sectional radii.

The circumference, or outer radius, of each layer was calculated based on the assumption that the particle is spherical and its cross section circular. The radius from machine learning was then compared to the radius from manual labeling according to Eq.4:

$$Difference\ (\%) = \frac{r_{ML} - r_m}{r_m} \times 100 \quad (4)$$

where $r_{ML}$ is the cross-section radius based on machine-learning segmentation and $r_m$ is the cross-section radius from manual labeling.

The majority (2,099 out of 2,171) of the calculated kernel radii from machine-learning segmentation are within ±5% of the manual results, as shown in Figure 12(a). A total of 19 images fall outside a range of the ±10% range. As shown in Figure 12(b), small kernel cross-section radii are the primary cause of large differences between the radii calculated from machine learning and those from manual labeling. This issue is evident in the small kernel shown in Figure 7(b); RU-Net 512 struggled to accurately segment it. In addition, when the cross section is far from the particle midplane, the shape of the kernel at the cross section may deviate significantly from that of a circle, contributing to a larger kernel difference in the calculated radii. For instance, the kernel in the cross section shown in Figure 13 is both small (with a radius of 47 μm compared to 213 μm at the mid-plane in the as-fabricated state [29]) and deviates from the circular shape. The IoU of the kernel prediction from machine learning is 70%, with the kernel difference being about -41%. Except for these extreme cases, overall the cross-section radii from machine learning match well with those from manual annotation.

Furthermore, the distribution of kernel difference values in Figure 12[a] is skewed in the negative direction, indicating in a lot of cases, the machine learning approach yields smaller radii than the manual approach. The $r_{ML}$ is derived from a circle with the same area as the segmented kernels from machine learning, while manual annotation connects 100 user-selected points to determine the perimeter and calculate the radius from the perimeter. Irregularities in the kernel shape and roughness on the kernel surface can result in a larger radius compared to the implied circle with a smooth surface. Although the overestimation from manual annotation is small in most cases in this dataset [11], it inevitably results in a negatively skewed kernel difference distribution. This pattern is consistent across all layers, as shown below.

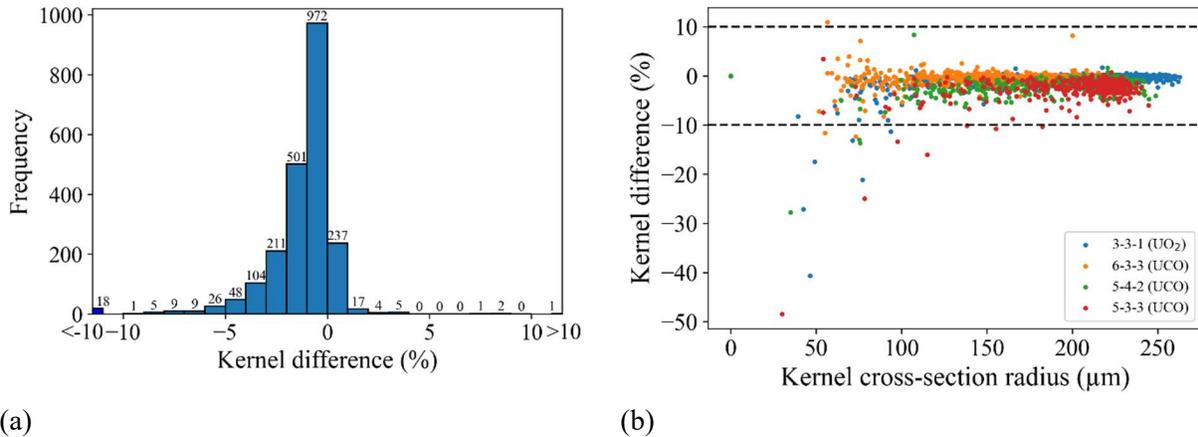

(a)                                                                 (b)

Figure 12. (a) A histogram of the kernel cross-section radius difference between manual labeling (ground truth) and machine learning. (b) Kernel cross-section radius difference as a function of the kernel cross section radius from the manual labeling.

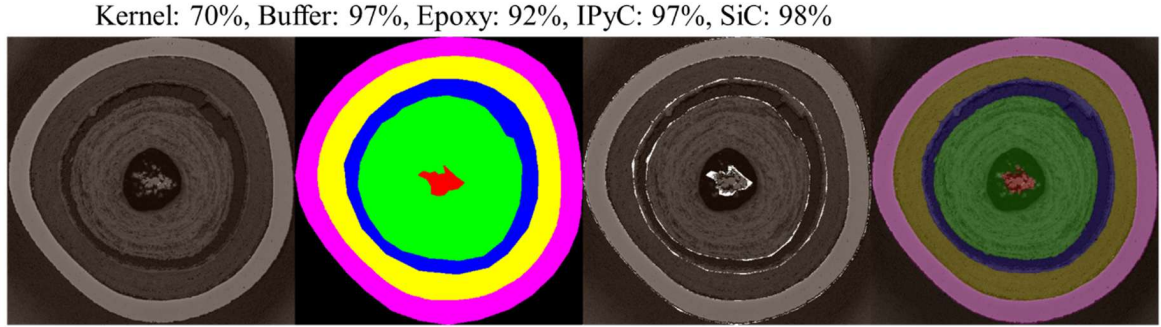

Figure 13. An example of a cross section with a small and irregular kernel. *Left to right*: input image, ground truth, input image overlaid with the difference (shown in white) between the ground truth and prediction, and input image overlaid with the prediction. The IoU of prediction for each layer is kernel 70%, buffer 97%, epoxy 92%, and SiC 98%.

Figure 14 shows that the outer buffer radii from machine learning closely match those from manual labeling, with only four instances falling outside the ±10% range. The worst case, with a difference of -13%, is presented in Figure 15. The layer segmentation for this cross section is generally accurate; the IoU for the buffer is 96%. However, the buffer misses a piece at the three o'clock position, and a fragment of buffer bridges to the IPyC at the six o'clock position, distorting its shape from a perfect circular ring. Despite these minor discrepancies in the worst case, the outer buffer radii from machine learning generally agree with with those from manual annotation.

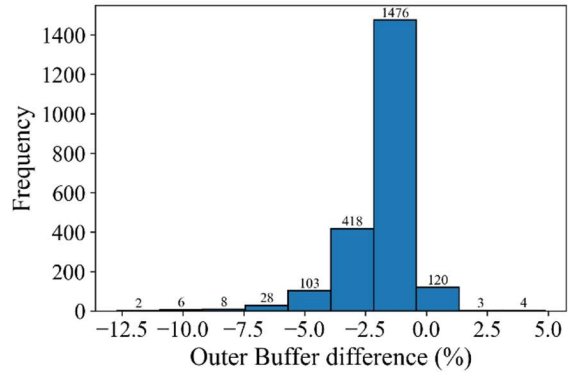

Figure 14. A histogram of the outer buffer radius difference between manual labeling (ground truth) and machine learning.

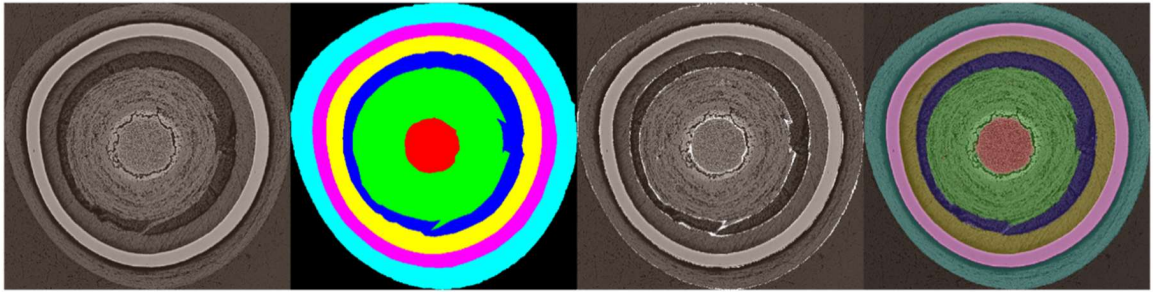

Figure 15. The cross section with the largest difference (-13%) in buffer radius. *Left to right*: input image, ground truth, input image overlaid with the difference (shown in white) between the ground truth and prediction, and input image overlaid with the prediction. The IoU of prediction for each layer is kernel 70%, buffer 96%, epoxy 90%, SiC 96%, and OPyC 96%.

The radii of the inner IPyC and outer SiC obtained from machine-learning segmentation closely match those obtained from the manual annotation, as shown in Figure 16. This is not surprising, given the high IoU values for the IPyC and SiC, and the fact that these two layers are generally shaped like circular rings.

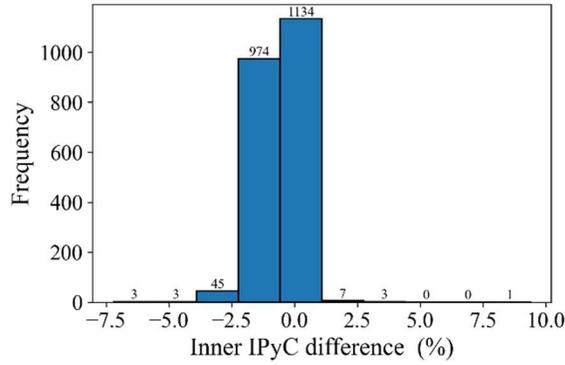 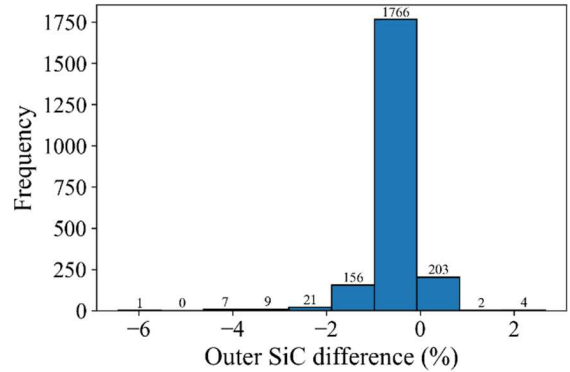

(a) (b)

Figure 16. A histogram of the radius difference between manual labeling (ground truth) and machine learning: (a) inner IPyC radii (b) outer SiC radii.

### 3.4 Sphere Radius Comparison—RU-Net 512

This section compares spherical radii between those derived from segmentation results of machine learning and those from the manual annotations. The spherical radii are determined through maximum likelihood fitting of cross-sectional radii at four cross sections of each particle. For spherical radii based on segmentation results from machine learning, maximum likelihood fitting was performed using initial five layers—kernel radius, buffer outer radius, IPyC inner radius, IPyC outer radius (or SiC inner radius), and SiC outer radius at four cross sections—for particles from all compacts except Compact 5-4-2. Subsequently, six layers are fitted for particles from Compact 5-4-2, including the OPyC layer. Note that the IPyC outer radius / SiC inner radius and OPyC outer radius are not included in the manual annotation. As stated in section 2.3, although there were 2,171 cross-sectional images, with 407 containing the OPyC layer, only 391 particles yield spherical radii through maximum likelihood fitting, of which 70 particles have OPyC radii. This is due to the following reasons: 1) some particles do not have all four workable cross sections, and 2) some optimization attempts fail to converge to a solution. The machine-learning spherical radii of the kernel, outer buffer, inner IPyC, and outer SiC were compared with those from manual annotation. As expected, the machine-learning SiC outer radius shows the best match with manual labeling, falling within -3.4% to +0.5% range, with the majority (329 out of 391) in the ±0.5% range (Figure 17[a]). The machine-learning IPyC inner radius has a good match with that of manual annotation, falling within a -5.4% to +3.95% range with only one instance outside the ±5% range (Figure 17 [b]). The machine-learning buffer outer radius has a range of -7.6% to +5.3% difference compared to manual labeling, with 12 instances outside the ±5% range (Figure 17 [c]). The kernel radius had the largest deviation, with a range of -8.7% to +12.0%, and 29 instances outside the ±5% range (Figure 17 [d]).

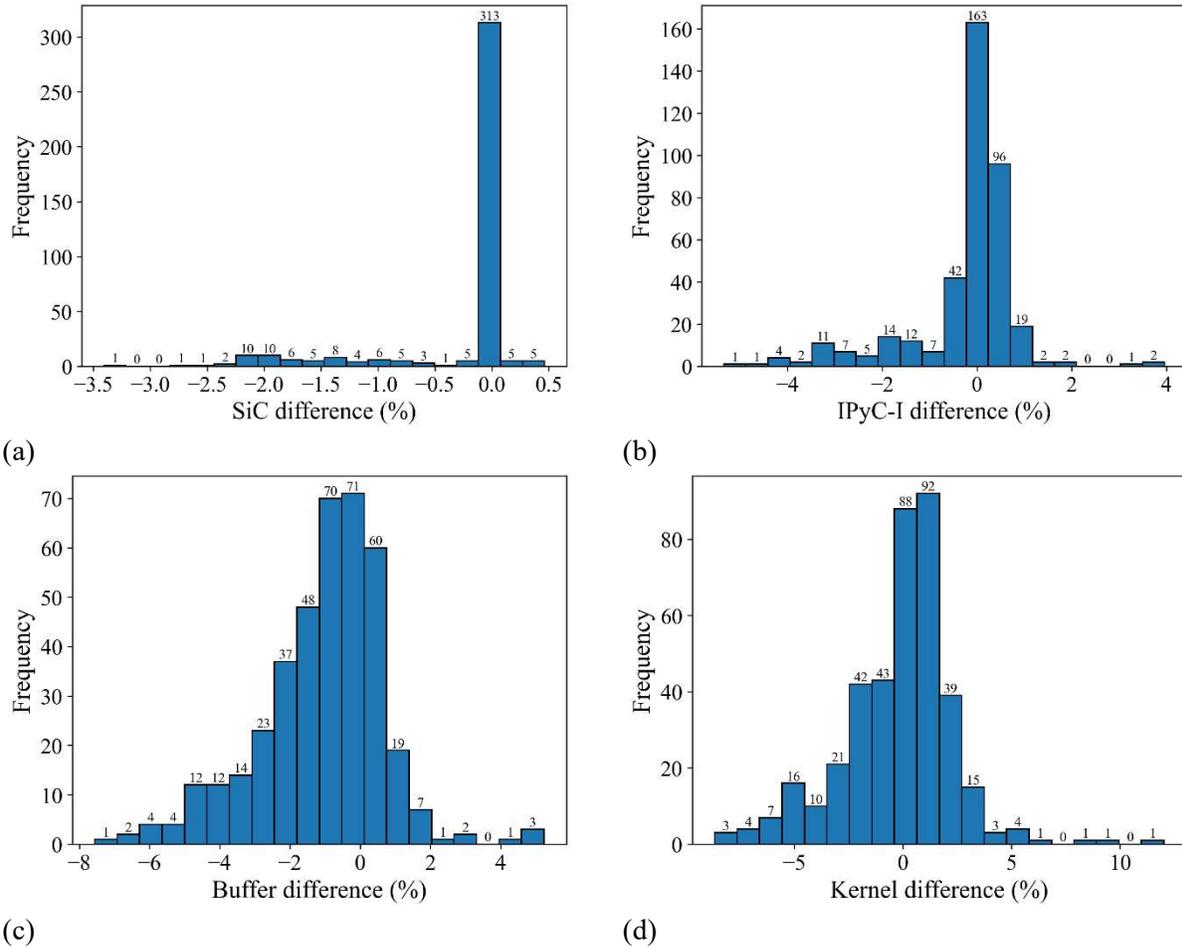

Figure 17. A histogram of spherical radius difference between manual annotation (ground truth) and machine learning: (a) SiC spherical radii, (b) inner IPyC spherical radii, (c) outer buffer spherical radii, and (d) kernel spherical radii.

The mean and standard deviation of the four layers from machine learning were compared to those from manual annotation, as presented in Figure 18(a) through 17(d). The as-fabricated mean radii and standard deviations were also shown as a line and shadow band; however, it is important to clarify in the fabrication report [29], only kernel radii and layer thickness are available. These mean radii and standard deviations for TRISO layers are calculated by summing the kernel radius and the layer thicknesses, while propagating the associated uncertainties. The mean values from both machine learning and manual annotation exhibit only a slight shift and have largely overlapping standard deviations, indicating a good match. The radii obtained from machine learning confirmed conclusions based on manual annotation, including obvious kernel swelling and buffer shrinkage. For Compact 3-3-1 ($UO_2$ kernel), the post-irradiated SiC layer outer radii were consistent with the as-fabricated ones, while for compacts with a UCO kernel, the SiC outer radii appeared slightly larger than the as-fabricated values. Similarly, when comparing the outer IPyC / inner SiC radii to the as-fabricated state (Figure 18[e]), Compact 3-3-1 displayed almost no change, whereas slight differences were observed for compacts with UCO kernels. However, these increase (4 µm in the mean SiC outer radius and 6.9 µm in the mean outer IPyC / inner SiC radius) are likely trivial, given the challenges in boundary identification, even with manual annotation. The manual annotation (ground truth) has a repeatability of ±3 µm [11]. In the input image resized to 512 by 512, where each pixel corresponds to approximately 1 - 2 µm in length, a one-pixel deviation in boundary prediction can result in a 1-2 µm discrepancy. Additionally, the change in the volume of the SiC is statistically insignificant. Therefore, it is concluded that there is no significant change in the SiC outer and inner radii.

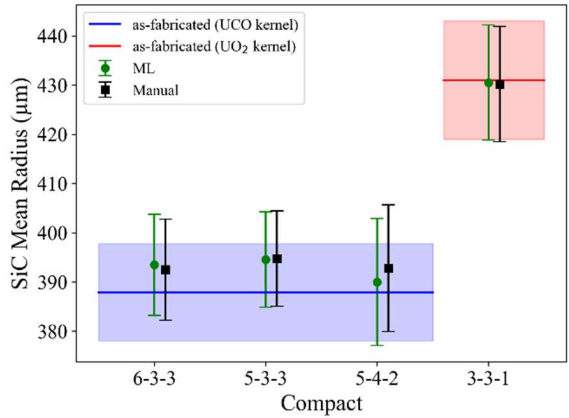

(a)

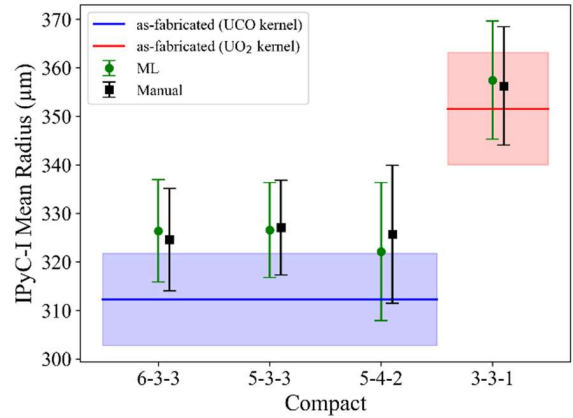

(b)

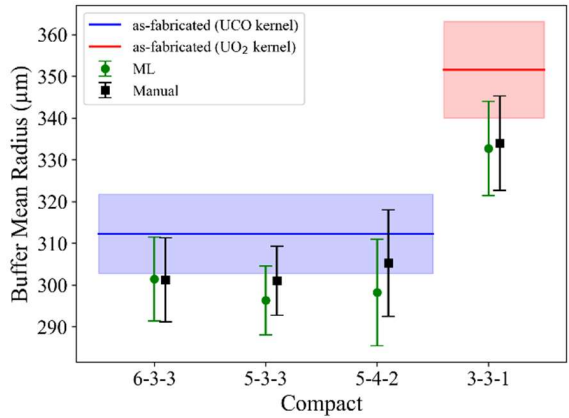

(c)

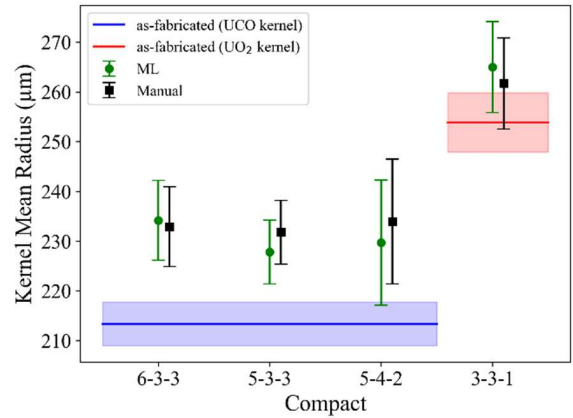

(d)

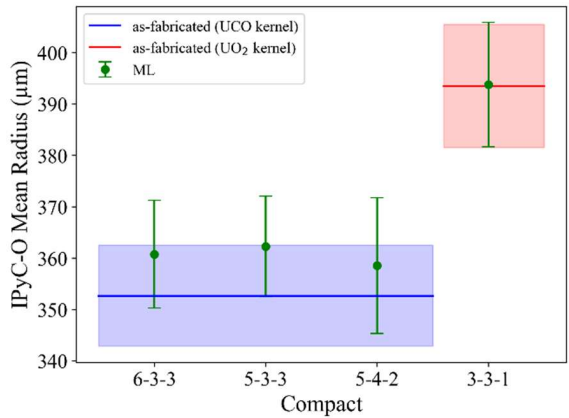

(e)

Figure 18. Comparison of mean and standard deviation of spherical radii among the as-fabricated state, machine-learning predictions, and manual annotations: (a) outer SiC, (b) inner IPyC, (c) outer buffer, (d) kernel, and (e) outer IPyC (or inner SiC).

For Compact 5-4-2, maximum likelihood fitting was performed using six layers, including an additional OPyC layer. The histogram of observed OPyC radius after irradiation was compared to the as-fabricated radius assuming a normal distribution (Figure 19). The as-fabricated OPyC layer radius was 431.25 ± 10.30 μm [29]. Based on 70 particles from Compact 5-4-2, the resulting irradiated OPyC radius was 430.57 ± 11.67 μm. The mean radius decreased slightly, but this change is statistically insignificant.

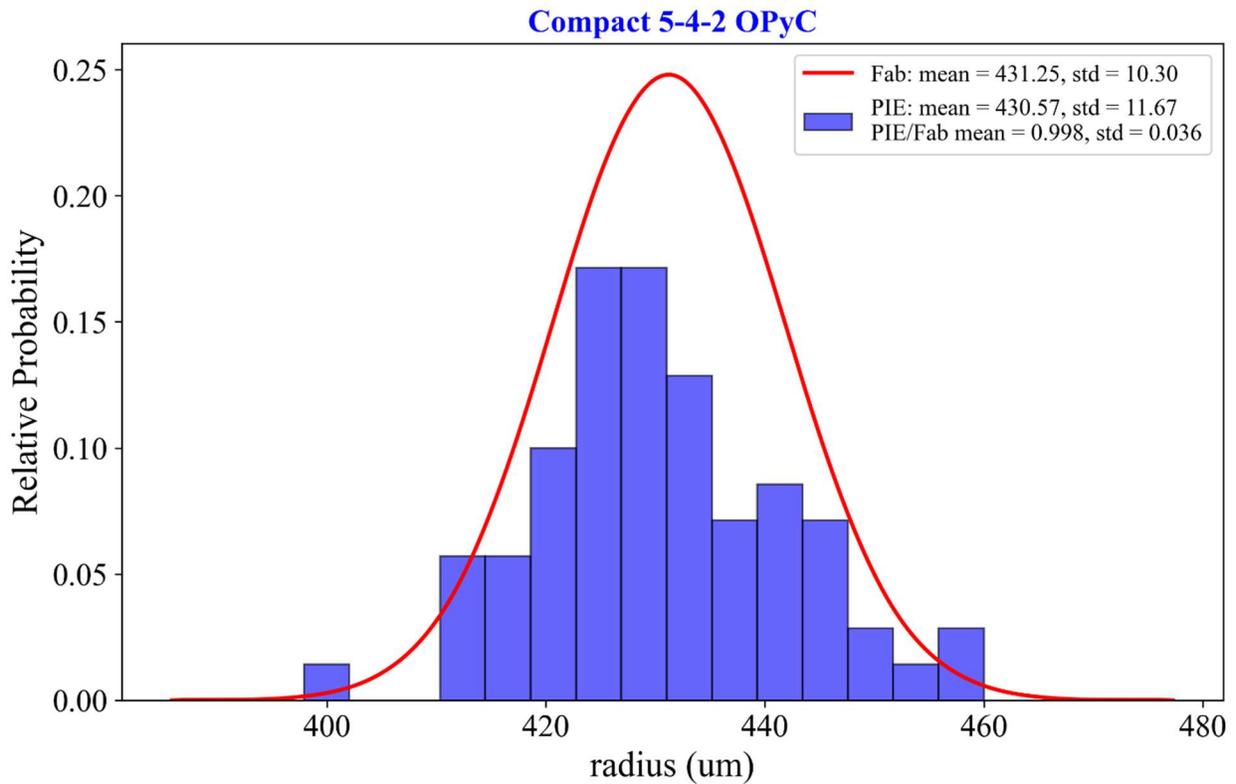

Figure 19. A histogram of OPyC spherical radius from Compact 5-4-2 for the as-fabricated and post-irradiation states. The post-irradiation / as-fabricated ratio of the mean radius is 0.999 ± 0.035.

## 4. CONCLUSION

A comprehensive dataset composed of 2,171 cross-section images with corresponding manual annotations was generated for machine-learning training. We subsequently developed RU-Net, a novel neural network architecture that integrates one original encoder with ResNet encoders to enhance the model's ability to delineate boundaries between different layers with greater clarity. We compared RU-Net with several widely recognized CNNs, including U-Net, ResNet, and Attention U-Net, and it demonstrated superior performance, achieving an impressive IoU exceeding 93% for both the training and test datasets.

This evaluation was conducted using the mean Intersection over Union (mIoU) metric, a standard measure for assessing the accuracy of segmentation models. Specifically, RU-Net performed best in the following areas: (a) recognizing the absence of kernels in the unbalanced dataset (only eight images without kernels), (b) identifying abnormalities due to sample preparation damages, such as excess epoxy over IPyC and broken OPyC. However, it faced challenges with SiC that had polishing scratches, sometimes misidentifying scratches as OPyC in particles without OPyC layer.

When comparing RU-Net with two input image resolutions (512 by 512 pixels and 256 by 256 pixels), RU-Net 256's performance in identifying classes of IPyC, SiC, OPyC, and background was comparable to that of RU-Net 512. However, it did not perform as well in identifying kernels, buffers, and epoxy classes. This is reasonable as the reduced image resolution decreases the boundary clarity between layers. On the positive side, RU-Net 256 can be trained on a typical computer without requiring GPU resources.

The cross-section radii and fitted spherical radii obtained from RU-Net 512 and manual annotation were compared and showed a reasonably good match. In addition, the spherical radii of outer IPyC / inner SiC as well as of the OPyC were determined. There were no significant changes in the outer IPyC / inner SiC and OPyC outer radii.

In conclusion, RU-Net has successfully segmented TRISO layers. This automatic segmentation can significantly reduce the subjectivity and manual effort required for the statistical analysis of particles. In addition to layer radii, other information such as the gap size and the roundness of layers can be easily derived from this segmentation. Ultimately, RU-Net paves the way for detailed, automatic microstructure analysis of large sample sizes needed for researching TRISO-coated fuel particle microspheres.

**Acknowledgments**

The authors acknowledge the financial support from the U.S. Department of Energy, Advanced Fuels Campaign (AFC) of the Nuclear Technology Research and Development program in the Office of Nuclear Energy. This research made use of the resources of the High-Performance Computing Center (HPC) at Idaho National Laboratory (INL), which is supported by the Office of Nuclear Energy of the U.S. Department of Energy and the Nuclear Science User Facilities (NSUF). This manuscript has been authorized by Battelle Energy Alliance, LLC, under Contract No. DE-AC07–05ID14517 with the U.S. Department of Energy. And the authors greatly acknowledge all Materials and Fuels Complex (MFC) facility support regarding sample preparation and handling, and data collection.

**Data availability statement**

The data that support the findings of this study are available from the corresponding author upon reasonable request.